\def\year{2021}\relax
\documentclass[letterpaper]{article} 
\usepackage{aaai21}  
\usepackage{times}  
\usepackage{helvet} 
\usepackage{courier}  
\usepackage[hyphens]{url}  
\usepackage{graphicx} 
\usepackage{natbib}  
\usepackage{caption} 
\usepackage{amsmath}
\usepackage{slashbox}   
\usepackage{subcaption}
\usepackage{tabularx}
\usepackage{booktabs}
\usepackage{verbatim}
\usepackage{hhline}
\usepackage{amssymb}
\usepackage{amsthm}
\usepackage{rotating}

\newcommand{\Tref}[1]{Table~\ref{#1}}
\newcommand{\Eref}[1]{Eq.~(\ref{#1})}

\renewcommand{\paragraph}[1]{\noindent{\bf #1.}}
\newtheorem{prop}{Proposition}
 
\DeclareMathOperator*{\argmax}{arg\,max}

\title{Improving Gradient Flow with Unrolled Highway Expectation Maximization}
\author{
    Chonghyuk Song, Eunseok Kim, Inwook Shim\thanks{Inwook Shim is the corresponding author.}\\
}
\affiliations{
    Ground Technology Research Institute, Agency for Defense Development
}

\begin{document}

\maketitle

\begin{abstract}
Integrating model-based machine learning methods into deep neural architectures allows one to leverage both the expressive power of deep neural nets and the ability of model-based methods to incorporate domain-specific knowledge. In particular, many works have employed the expectation maximization (EM) algorithm in the form of an unrolled layer-wise structure that is jointly trained with a backbone neural network. However, it is difficult to discriminatively train the backbone network by backpropagating through the EM iterations as they are prone to the vanishing gradient problem. To address this issue, we propose Highway Expectation Maximization Networks~(HEMNet), which is comprised of unrolled iterations of the generalized EM~(GEM) algorithm based on the Newton-Rahpson method. HEMNet features scaled skip connections, or highways, along the depths of the unrolled architecture, resulting in improved gradient flow during backpropagation while incurring negligible additional computation and memory costs compared to standard unrolled EM. Furthermore, HEMNet preserves the underlying EM procedure, thereby fully retaining the convergence properties of the original EM algorithm. We achieve significant improvement in performance on several semantic segmentation benchmarks and empirically show that HEMNet effectively alleviates gradient decay.
\end{abstract}

\section{Introduction}

The Expectation Maximization~(EM) algorithm \cite{Dempster77maximumlikelihood} is a well-established algorithm in the field of statistical learning used to iteratively find the maximum likelihood solution for latent variable models. It has traditionally been used for a variety of problems, ranging from unsupervised clustering to missing data imputation.
With the dramatic rise in adoption of deep learning in the past few years, recent works~\cite{jampani18ssn, emrouting, li19emanet, wang_2020, nem} have aimed to combine the model-based approach of EM with deep neural networks. These two approaches are typically combined by unrolling the iterative steps of the EM as layers in a deep network, which takes as input the features generated by a backbone network that learns the representations. Jointly training this combined architecture discriminatively allows one to leverage the expressive power of deep neural networks and the ability of model-based methods to incorporate prior knowledge of the task at hand, resulting in potentially many benefits~\cite{deep_unfolding}.

First, unrolled EM iterations are analogous to an attention mechanism when the underlying latent variable model is a Gaussian Mixture Model (GMM) \cite{emrouting, li19emanet}. The Gaussian mean estimates capture long-range interactions among the inputs, just like in the self-attention mechanism \cite{transformer, NonLocal2018, zhao2018psanet}. Furthermore, EM attention \cite{li19emanet} is computationally more efficient than the original self-attention mechanism, which computes representations as a weighted sum of every point in the input, whereas EM attention computes them as a weighted sum of a smaller number of Gaussian means. The EM algorithm iteratively refines these Gaussian means such that they monotonically increase the log likelihood of the input, increasingly enabling them to reconstruct and compute useful representations of the original input.

Despite the beneficial effects EM has on the forward pass, jointly training a backbone neural network with EM layers is challenging as they are prone to the vanishing gradient problem \cite{li19emanet}. This phenomenon, which was first introduced in \cite{transformer}, is a problem shared by all attention mechanisms that employ the dot-product softmax operation in the computation of attention maps. Skip connections have shown to be remarkably effective at resolving vanishing gradients for a variety of deep network architectures \cite{highway_nets, identity_mappings_in_resnets, densenet, lstm_with_forget_gate, lstm, gru}, including attention-based models \cite{transformer, transparentatt, deeptransformer}. However, the question remains as to \textit{how} to incorporate skip connections in way that maintains the underlying EM procedure of monotonically converging to a (local) optimum of the data log-likelihood.


In this paper, we aim to address the vanishing gradient problem of unrolled EM iterations while preserving the EM algorithm, thereby retaining its efficiency and convergence properties and the benefits of end-to-end learning. Instead of unrolling EM iterations, we unroll \textit{generalized} EM (GEM) iterations, where the M-step is replaced by one step of the Newton-Rahpson method \cite{gem_newton}. This is motivated by the key insight that unrolling GEM iterations introduces weighted skip connections, or highways \cite{highway_nets}, along the depths of the unrolled architecture, thereby improving its gradient flow during backpropgation. The use of Newton's method is non-trivial. Not only do GEM iterations based on Newton's method require minimal additional computation and memory costs compared to the original EM, but they are also guaranteed to improve the data log-likelihood. To demonstrate the effectiveness of our approach, we formulate the proposed GEM iterations as an attention module, which we refer to as Highway Expectation Maximization Network (HEMNet), for existing backbone networks and evaluate its performance on challenging semantic segmentation benchmarks.

\section{Related Works}
\paragraph{Unrolled Expectation Maximization} 
With the recent rise in adoption of deep learning, many works have incorporated modern neural networks with the well-studied EM algorithm to leverage its clustering and filtering capabilities. SSN \cite{jampani18ssn} combine unrolled EM iterations with a neural network to learn task-specific superpixels. CapsNet \cite{emrouting} use unrolled EM iterations as an attentional routing mechanism between adjacent layers of the network. EMANet \cite{li19emanet} designs an EM-based attention module that boosts the performance of a backbone network on semantic segmentation. A similar module is used in \cite{wang_2020} as a denoising filter for fine-grained image classification. On the other hand, NEM \cite{nem} incorporates the \textit{generalized} EM (GEM) algorithm \cite{wu1983} to learn representations for unsupervised clustering, where the M-step of the original EM algorithm is replaced with one gradient ascent step towards improving the data log-likelihood. Unlike our proposed method and original EM, NEM does not guarantee an improvement of the data log-likelihood.

\paragraph{Skip Connections} Skip connections are direct connections between nodes of different layers of a neural network that bypass, or skip, the intermediate layers. They help overcome the vanishing gradient problem associated with training very deep neural architectures \cite{vanishinggradients} by allowing gradient signals to be directly backpropagated between adjacent layers \cite{resnet, highway_nets, densenet, lstm, gru}. In particular, skip connections 
are crucial for training attention-based models, which are also prone to vanishing gradients \cite{transparentatt, deeptransformer, improving_deep_transformer}. The Transformer \cite{transformer} employs an identity skip connection around each of the sub-layers of the network, without which the training procedure collapses, resulting in significantly worse performance \cite{transparentatt}. Subsequent works \cite{transparentatt, deeptransformer} were able to train deeper Transformer models by creating weighted skip connections along the depth of the encoder of the Transformer, providing multiple backpropagation paths and improving gradient flow.

Our approach is motivated by the success of the above works in combating vanishing gradients and in fact structurally resembles Highway Networks \cite{highway_nets} and Gated Recurrent Units \cite{gru}. The key difference is that our method introduces skip connections into the network in a way that preserves the underlying EM procedure and by extension its convergence properties and computational efficiency.


\begin{figure*}[t!]
  \includegraphics[width = 0.70\textwidth,trim={0 3.5mm 0 0},clip]{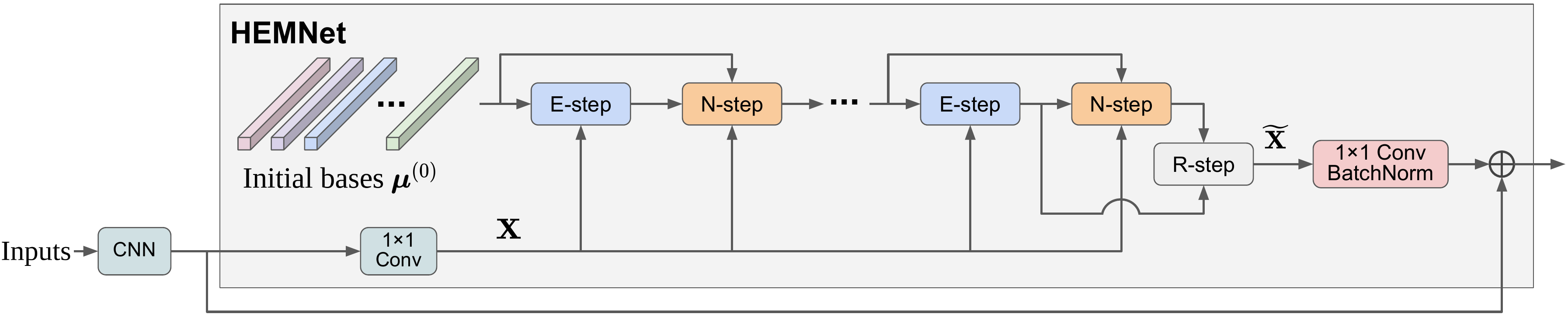}
  \centering
  \caption{High-level structure of the proposed HEMNet}
  \label{fig:highlevel_hemnet}
\end{figure*}

\section{Preliminaries}
\subsection{EM Algorithm for Gaussian Mixture Models}
\label{em-gmm}
The EM algorithm is an iterative procedure that finds the maximum likelihood solution for latent variable models. A latent variable model is described by the joint distribution $p(\mathbf{X}, \mathbf{Z} | \boldsymbol{\theta})$, where $\mathbf{X}$ and $\mathbf{Z}$  denote the dataset of $N$ observed samples $\mathbf{x}_{n}$ and the corresponding set of latent variables $\mathbf{z}_{n}$, respectively, and $\boldsymbol{\theta}$ denotes the model parameters. The Gaussian mixture model (GMM) \cite{gmm} is a widely used latent variable model that models the distribution of observed data point $\mathbf{x}_{n}$ as a linear superposition of $K$ Gaussians:
\begin{equation}
\label{eq:gmm_def}
p(\mathbf{x}_{n}) = \sum_{k=1}^{K}\boldsymbol{\pi}_{k}\mathcal{N}(\mathbf{x}_{n}|\boldsymbol{\mu}_{k},\boldsymbol{\Sigma}_{k}),
\end{equation}
\begin{equation}
\label{eq:log_likelihood}
\ln p(\mathbf{X}|\boldsymbol{\theta}) = \sum_{n=1}^{N}\ln \left\{\sum_{k=1}^{K}\mathcal{N}(\mathbf{x}_{n}|\boldsymbol{\mu}_{k},\boldsymbol{\Sigma}_{k})\right\},
\end{equation}\\
where the mixing coefficient $\boldsymbol{\pi}_{k}$, mean $\boldsymbol{\mu}_{k}$, and covariance $\boldsymbol{\Sigma}_{k}$ constitute the parameters for the $k^{th}$ Gaussian.
We use a fixed isotropic covariance matrix $\boldsymbol{\Sigma}_{k} = \sigma^{2}\mathbf{I}$ and drop the mixing coefficients as done in many real applications. The EM algorithm aims to maximize the resulting log-likelihood function in \Eref{eq:log_likelihood} by performing coordinate ascent on its evidence lower bound $\mathcal{L}$ (ELBO) \cite{coordinatedescent}:
\begin{align}
    \begin{split}
         \ln p(\mathbf{X}|\boldsymbol{\theta}) & \geq \underbrace{\sum_{\mathbf{Z}}q(\mathbf{Z})\ln p(\mathbf{X}, \mathbf{Z}|\boldsymbol{\theta})}_{Q(\boldsymbol{\theta}, q(\mathbf{Z}))} + \underbrace{\sum_{\mathbf{Z}}-q(\mathbf{Z})\ln q(\mathbf{Z})}_{H(q(\mathbf{Z}))} \\
         \label{eq:elbo_gmm}
         & = \underbrace{\sum_{n=1}^{N}\sum_{k=1}^{K}\gamma_{nk}\ln\mathcal{N}(\mathbf{x_{n}}|\boldsymbol{\mu_{k}},\boldsymbol{\Sigma_{k}})-\gamma_{nk}\ln\gamma_{nk}}_{\mathcal{L}(\boldsymbol{\mu, \gamma})},
    \end{split}
\end{align}
which is the sum of the expected complete log-likelihood $Q(\boldsymbol{\theta}, q)$ and entropy term $H(q)$, for any arbitrary distribution $q(\mathbf{Z})$ defined by $\sum_{k}\gamma_{nk} = 1$. By alternately maximizing the ELBO with respect to $q$ and $\boldsymbol{\theta}$ via the E- and M-step, respectively, the log-likelihood is monotonically increased in the process and converges to a (local) optimum: 
\begin{align}
\label{eq:estep_gmm}
\textrm{\textbf{E-step}:}&\quad
\gamma_{nk} = \dfrac{\mathcal{N}(\mathbf{x}_{n}|\boldsymbol{\mu}^{old}_{k},\boldsymbol{\Sigma}_{k})}{\sum_{j=1}^{K}\mathcal{N}(\mathbf{x}_{n}|\boldsymbol{\mu}^{old}_{j},\boldsymbol{\Sigma}_{j})}\\
\label{eq:mstep_gmm}
\textrm{\textbf{M-step}:}&\quad
\boldsymbol{\mu}_{k}^{new} = \dfrac{1}{N_{k}}\sum_{n=1}^{N}\gamma_{nk}\mathbf{x}_{n}
\end{align}

In the E-step, the optimal $q$ is $p(\mathbf{Z}|\mathbf{X}, \boldsymbol{\theta}^{old})$, the posterior distribution of $\mathbf{Z}$. \Eref{eq:estep_gmm} shows the posterior for GMMs, which is described by $\gamma_{nk}$, the responsibility that the $k^{th}$ Gaussian basis $\boldsymbol{\mu}_{k}$ takes for explaining observation $\mathbf{x}_{n}$.
In the M-step, the optimal $\boldsymbol{\theta}^{new}=  \argmax_{\boldsymbol{\theta}}Q(\boldsymbol{\theta}, q(\boldsymbol{\theta}^{old}))$ since the entropy term of the ELBO isn't dependent on $\boldsymbol{\theta}$. For GMMs, this $\argmax$ is tractable, resulting in the closed-form of \Eref{eq:mstep_gmm}, where $N_{k} = \sum_{n=1}^{N}\gamma_{nk}$.

When the M-step is \textit{not} tractable however, we resort to the \textit{generalized} EM (GEM) algorithm, whereby instead of maximizing $Q(\boldsymbol{\theta}, q)$ with respect to $\boldsymbol{\theta}$ the aim is to at least increase it, typically by taking a step of a nonlinear optimization method such as gradient ascent or the Newton-Raphson method. In this paper, we use the GEM algorithm based on the Newton-Raphson method for its favorable properties, which are described in section 4.2:
\begin{align}
\label{eq:gem_general_form_append}
\boldsymbol{\theta}^{new} &= \boldsymbol{\theta}^{old} - \eta\left[ \left({\dfrac{\partial^{2}Q}{\partial\boldsymbol{\theta}\partial\boldsymbol{\theta}}}\right)^{-1}\dfrac{\partial Q}{\partial\boldsymbol{\theta}}\right]_{\boldsymbol{\theta}=\boldsymbol{\theta}^{old}}
\end{align}


\subsection{Unrolled Expectation Maximization}
In this section, we no longer consider EM iterations as an algorithm, but rather as a sequence of layers in a neural network-like architecture. This structure, which we refer to as ``unrolled EM'', is comprised of a pre-defined $T$ number of alternating E- and M-steps, both of which are considered as network layers that take as input the output of its previous step and the feature map $\mathbf{X}$ generated by a backbone CNN. For simplicity, we consider the feature map  $\mathbf{X}$ of shape $C \times H \times W$ from a single sample, which we reshape into $N \times C$, where $N = H \times W$.

Given the CNN features $\mathbf{X} \in \mathbb{R}^{N \times C}$ and Gaussian bases from the \textit{t}-th iteration $\boldsymbol{\mu}^{(t)} \in \mathbb{R}^{K \times C}$, the E-step computes the responsibilities $\boldsymbol{\gamma}^{(t+1)} \in \mathbb{R}^{N \times K}$ according to \Eref{eq:estep_gmm}, which can be rewritten in terms of the RBF kernel $\exp (-||\mathbf{x}_{n} - \boldsymbol{\mu}_{k}||^{2}_{2} / \sigma^{2})$:
\begin{equation}
    \gamma^{(t+1)}_{nk} = \dfrac{\exp (-||\mathbf{x}_{n} - \boldsymbol{\mu}^{(t)}_{k}||^{2}_{2} / \sigma^{2})}{\sum_{j=1}^{K}\exp (-||\mathbf{x}_{n} - \boldsymbol{\mu}^{(t)}_{j}||^{2}_{2} / \sigma^{2})}
\end{equation}
\begin{equation}
    \label{eq:dot_product_softmax}
    \textrm{\textbf{E-step}:}\quad\boldsymbol{\gamma}^{(t+1)}= \textrm{softmax}\left(\dfrac{\mathbf{X}\boldsymbol{\mu}^{(t)\top}}{\sigma^{2}}\right)
\end{equation}

As shown in \Eref{eq:dot_product_softmax}, the RBF kernel can be replaced by the exponential inner dot product $\exp(\mathbf{x}_{n}^{\top}\boldsymbol{\mu}_{k}/\sigma^{2})$, which brings little difference to the overall results~\cite{NonLocal2018, li19emanet} and can be efficiently implemented by a softmax applied to a matrix multiplication operation scaled by the temperature $\sigma^{2}$.

The M-step then updates the Gaussian bases according to \Eref{eq:mstep_gmm}, which is implemented by a matrix multiplication between normalized responsibilities $\bar{\boldsymbol{\gamma}}$ and features $\mathbf{X}$:
\begin{equation}
    \textrm{\textbf{M-step}:}\quad\boldsymbol{\mu}^{(t+1)}= \bar{\boldsymbol{\gamma}}^{(t+1)}\mathbf{X},
\end{equation}

After unrolling $T$ iterations of EM, the input $\mathbf{x}_{n}$, whose distribution was modeled as a mixture of Gaussians, is reconstructed as a weighted sum of the converged Gaussian bases, with weights given by the converged responsibilities \cite{li19emanet, wang_2020}. As a result, reconstructing the input features $\widetilde{\mathbf{X}} \in \mathbb{R}^{N \times C}$, which we call the R-step, is also implemented by matrix multiplication:
\begin{equation}
    \label{eq:rstep_matmul}
    \textrm{\textbf{R-step}:}\quad\widetilde{\mathbf{X}}= \boldsymbol{\gamma}^{(T)}\boldsymbol{\mu}^{(T)}
\end{equation}

Unrolling $T$ iterations of E- and M-steps followed by one R-step incurs $O(NKT)$ complexity \cite{li19emanet}. However, $T$ can be treated as a small constant for the values used in our experiments, resulting in a complexity of $O(NK)$.

\section{Highway Expectation Maximization Networks}
\subsection{Vanishing Gradient Problem of EM}
\label{subsection: vanishing gradient}
Vanishing gradients in unrolled EM layers stem from the E-step's scaled dot-product softmax operation, shown in \Eref{eq:dot_product_softmax}. This also happens to be the key operation of the self-attention mechanism in the Transformer \cite{transformer}, which was first proposed to address vanishing gradients associated with softmax saturation; without scaling, the magnitude of the dot-product logits grows larger with increasing number of channels $C$, resulting in a saturated softmax with extremely small local gradients. Therefore, gradients won't be backpropagated to layers below a saturated softmax. The Transformer counteracts this issue in self-attention layers by setting the softmax temperature $\sigma^{2} = \sqrt C$, thereby curbing the magnitude of the logits.


However, even with this dot-product scaling operation the Transformer is still prone to gradient vanishing, making it extremely difficult to train very deep models \cite{transparentatt, deeptransformer}. In fact, the training procedure has shown to even collapse when residual connections are removed from the Transformer \cite{transparentatt}. To make matters worse, an EM layer only has a single gradient path through $\boldsymbol{\mu}^{(t)}$ that reach lower EM layers, as opposed to the self-attention layer, which backpropagates gradients through multiple paths and therefore has shown to prevent more severe gradient vanishing \cite{improving_deep_transformer}.

\subsection{Unrolled Highway Expectation Maximization}
\label{subsection:unrolled_hem}
In order to resolve the vanishing gradient problem in unrolled EM layers, we propose Highway Expectation Maximization Networks~(HEMNet), which is comprised of unrolled GEM iterations based on the Newton-Raphson method, as shown in Fig.~\ref{fig:highlevel_hemnet}. The key difference between HEMNet and unrolled EM is that the original M-step is replaced by one Newton-Raphson step, or N-step:

\textbf{N-step:}
\begin{align}
\label{eq:gem_gmm}
\boldsymbol{\mu}_{k}^{(t+1)} &= \boldsymbol{\mu}_{k}^{(t)} - \eta\left(\dfrac{\partial^{2}\mathcal{Q}}{\partial\boldsymbol{\mu}_{k}\partial\boldsymbol{\mu}_{k}}\right)^{-1}\dfrac{\partial\mathcal{Q}}{\partial\boldsymbol{\mu}_{k}}
\\ &= \boldsymbol{\mu}^{(t)}_{k} - \eta\frac{-\sigma^{2}}{N_{k}^{(t+1)}} \left\{\sum_{n=1}^{N}\frac{\gamma_{nk}^{(t+1)}}{\sigma^{2}}\big(\mathbf{x}_{n} - \boldsymbol{\mu}^{(t)}_{k}\big)\right\}
\\ \label{eq:weighted_sum}
& = \underbrace{(1 - \eta)\boldsymbol{\mu}^{(t)}_{k}}_\textrm{skip connection} + \eta \underbrace{\left(\frac{1}{N_{k}^{(t+1)}} \sum_{n=1}^{N}\gamma_{nk}^{(t+1)}\mathbf{x}_{n}\right)}_\textrm{$\mathcal{F}^{EM}\big(\boldsymbol{\mu}_{k}^{(t)}, \mathbf{X}\big)$},
\end{align}
where $\eta$ is a hyperparameter that denotes the step size. For GMMs, the N-step update is given by \Eref{eq:gem_gmm}, which rearranged becomes \Eref{eq:weighted_sum}, a weighted sum of the current $\boldsymbol{\mu}_{k}^{(t)}$ and the output of one EM iteration $\mathcal{F}^{EM}\big(\boldsymbol{\mu}^{(t)}_{k}, \mathbf{X}\big)$. Interestingly, the original M-step is recovered when $\eta = 1$, implying that the N-step generalizes the EM algorithm.

\Eref{eq:weighted_sum} is significant as the first term introduces a skip connection that allows gradients to be directly backpropagated to earlier EM layers, thereby alleviating vanishing gradients. Furthermore, the weighted-sum update of the N-step endows HEMNet with two more crucial properties: improving the ELBO in the forward pass and incurring negligible additional space-time complexity compared to unrolled EM.

\begin{figure}[t]
  \includegraphics[width = \linewidth,trim={0 0.5mm 0 0},clip]{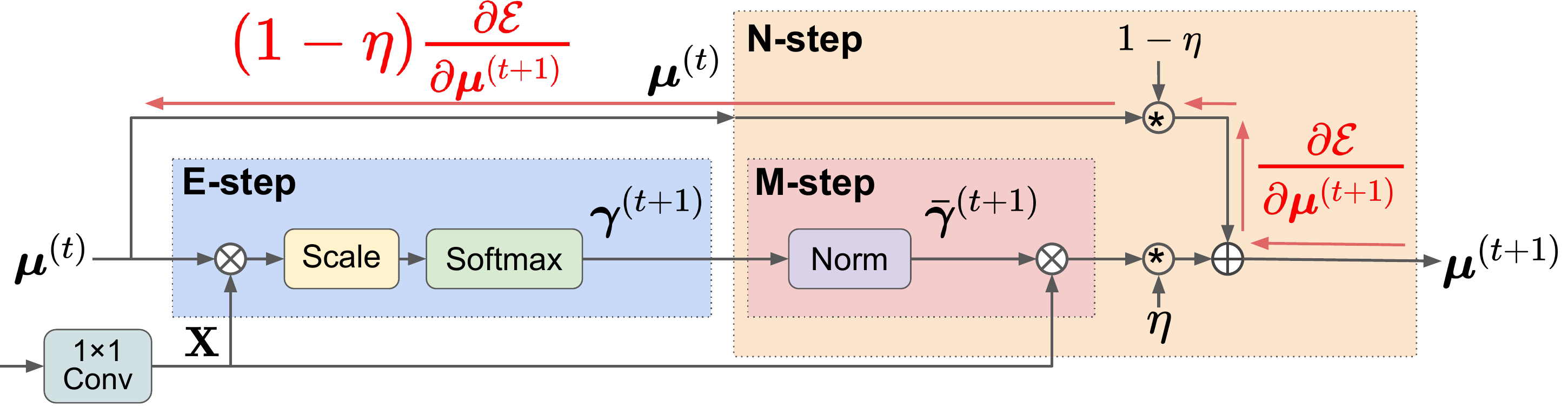}
  \centering
  \caption{
  Architecture of single HEM layer, which is comprised of one E-step and N-step operation
  }
  \label{fig:hemlayer}
\end{figure}
\begin{table*}
\centering
\begin{minipage}[ht!]{0.48\linewidth}
    \caption{Ablation study on training iteration number $T_{train}$ and step size $\eta_{train}$ on PASCAL VOC. The rightmost column denotes EMANet \cite{li19emanet}, where $\eta_{train} = 1.0$.}
    \label{table: ablation1}
    \centering
    \resizebox{0.9\textwidth}{!}{
    \begin{tabular}{cccccc}
     \toprule
     \backslashbox{$T_{train}$}{$\eta_{train}$} & 0.1     & 0.2      & 0.4      & 0.8     & EMANet (1.0) \\
     \midrule
     1 & 77.17 & 77.50 & 77.92 & 77.26 & 77.31 \\
     2 & 77.16 & 77.80 & 77.50 & \textbf{78.10} & \textbf{77.55} \\
     3 & 77.05 & 77.82 & 77.81 & 77.94 & 76.51 \\
     4 & 77.64 & 77.73 & \textbf{78.11} & 77.10 & 76.63 \\
     6 & 77.46 & \textbf{78.22} & 77.83 & 77.40 & 76.26 \\
     8 & 77.48 & 78.14 & 77.60 & 78.04 & 76.48 \\
     12 & \textbf{77.74} & 78.11 & 77.73 & 77.84 & 76.17 \\
     16 & 77.64 & 77.65 & 77.81 & 77.06 & 76.29 \\
     \bottomrule
    \end{tabular}
}
\end{minipage}\hfill
\begin{minipage}[ht!]{0.48\linewidth}
  \caption{Ablation study on evaluation iteration number $T_{eval}$. For each $\eta_{train}$, we perform ablations on the best $T_{train}$ (underlined). The best $T_{eval}$ is highlighted in bold.}
  \label{table: ablation2}
  \centering
  \resizebox{0.9\textwidth}{!}{
  \begin{tabular}{cccccc}
    \toprule
    \backslashbox{$T_{eval}$}{$\eta_{train}$} & 0.1     & 0.2      & 0.4      & 0.8     & EMANet (1.0) \\
    \midrule
    1 & 64.52 & 70.30 & 73.20 & 77.16 & 76.26 \\
    2 & 70.72 & 75.50 & 76.94 & \underline{78.10} & \underline{77.55} \\
    3 & 73.72 & 77.33 & 77.85 & 78.34 & \textbf{77.73} \\
    4 & 75.34 & 77.96 & \underline{78.11} & \textbf{78.42} & 77.70\\
    6 & 76.88 & \underline{\textbf{78.22}} & 78.24 & 78.37 & 77.50 \\
    8 & 77.47 & 78.13 & \textbf{78.24} & 78.30 & 77.37 \\
    12 & \underline{77.74} & 77.98 & 78.19 & 78.26 & 77.21 \\
    16 & \textbf{77.84} & 77.90 & 78.17 & 78.23 & 77.16 \\
    \bottomrule
  \end{tabular}
}
\end{minipage}
\end{table*}
\begin{figure*}[ht!]
  \includegraphics[width = 0.85\linewidth,trim={0 5mm 0 0},clip]{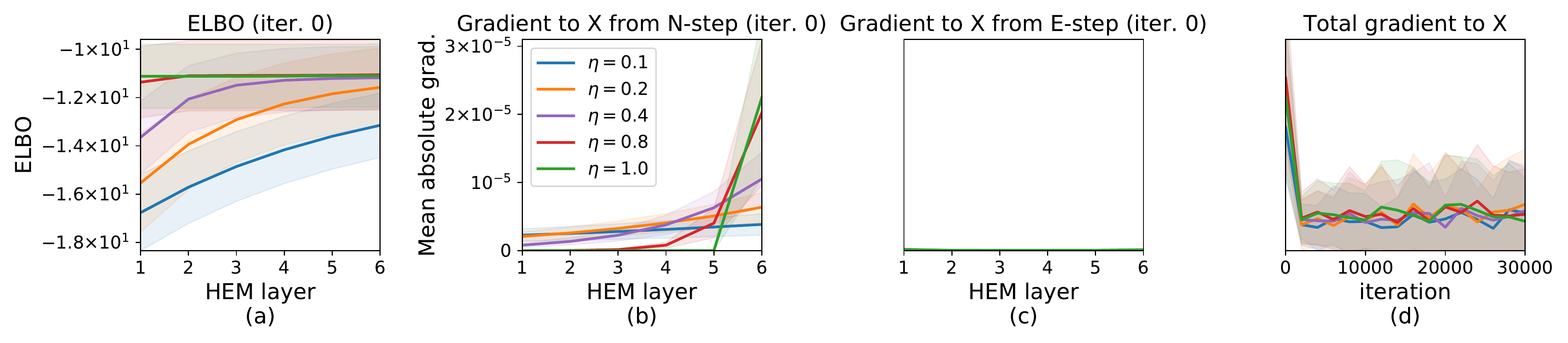}
  \centering
  \caption{The effects of changing $\eta_{train}$ on (a) the ELBO, (b) the gradient to input $\mathbf{X}$ from N-step and (c) from E-step at the beginning of training, and (d) the total gradient to $\mathbf{X}$ during training. The E-step gradients in (c) are on the order of $10^{-7}$ and is therefore not displayed. The mean absolute gradients were computed by backpropagating with respect to the same set of 100 randomly chosen training samples, every 2000 training iterations.}
  \label{fig:grad2x}
\end{figure*}
\subsubsection{Backward-pass properties: alleviating vanishing gradients}
The update equation of \Eref{eq:weighted_sum} resembles that of the Highway Network \cite{highway_nets}, which contain scaled skip connections, or highways, that facilitate information flow between neighboring layers. HEMNet also contains highways in between its unrolled iterations, or HEM layers, that bypass the gradient-attenuating E-step and allow gradients to be directly sent back to earlier HEM layers, as shown in Fig.~\ref{fig:hemlayer}. To show this, we derive expressions for the upstream gradients to each HEM layer and its (shared) input, by first recursively applying \Eref{eq:weighted_sum}:
\begin{equation}
\begin{aligned}
    \label{eq:nstep_recursively_applied}
    \boldsymbol{\mu}_{k}^{(T)} =& (1-\eta)^{(T-t)}\boldsymbol{\mu}_{k}^{(t)} \\&+ \left\{\sum_{i=t}^{T-1} \eta(1-\eta)^{(T-i-1)} \mathcal{F}^{EM}\big(\boldsymbol{\mu}_{k}^{(i)},\mathbf{X}\big)\right\},
\end{aligned}
\end{equation}
and then applying the chain rule to \Eref{eq:nstep_recursively_applied}, where $\mathcal{E}$, $T$, $\hat{\mathcal{F}}_{k}^{(i)}=\eta(1-\eta)^{(T-i-1)}\mathcal{F}^{EM}\big(\boldsymbol{\mu}_{k}^{(i)}, \mathbf{X}\big)$ are the loss function, number of HEM layers, and shorthand that absorbs the scalars, respectively:
\begin{equation}
\begin{split}
\label{eq:grad2mu}
\dfrac{\partial \mathcal{E}}{\partial\boldsymbol{\mu}^{(t)}_{k}} =& \dfrac{\partial \mathcal{E}}{\partial\boldsymbol{\mu}^{(T)}_{k}}\dfrac{\partial\boldsymbol{\mu}^{(T)}_{k}}{\partial\boldsymbol{\mu}^{(t)}_{k}}\\
=&\dfrac{\partial \mathcal{E}}{\partial\boldsymbol{\mu}^{(T)}_{k}}\bigg\{(1-\eta)^{(T-t)}+{\dfrac{\partial}{\partial\boldsymbol{\mu}^{(t)}_{k}}\sum_{i=t}^{T-1}\hat{\mathcal{F}}^{(i)}_{k}}\bigg\}
\end{split}
\end{equation}
It can be seen that the upstream gradients to $\boldsymbol{\mu}_{k}$ is the sum of two gradient terms: a term $(1-\eta)^{(T-t)}\frac{\partial \mathcal{E}}{\partial \boldsymbol{\mu}_{k}^{(T)}}$ directly propagated through the skip connections and a term $\frac{\partial \mathcal{E}}{\partial \boldsymbol{\mu}_{k}^{(T)}}\frac{\partial}{\partial\boldsymbol{\mu}^{(t)}_{k}}\sum_{i=t}^{T-1}\hat{\mathcal{F}}^{(i)}_{k}$ propagated through the E-step, which is negligible due to the E-step's vanishing gradient problem and hence can be ignored. This means that as we increase the scalar $(1-\eta)^{(T-t)}$ (by reducing $\eta$), the proportion of upstream gradients backpropagated to earlier HEM layers increases as well. Furthermore, it can be seen that when $\eta=1$ (the original EM case) the skip connection term in \Eref{eq:grad2mu} vanishes and leaves only gradients propagated through the gradient-attenuating E-step, highlighting the vanishing gradient problem of unrolled EM.

One consequence of \Eref{eq:grad2mu} is that as $\eta$ decreases, the backbone network parameters will be increasingly influenced by earlier HEM layers, as shown in the following derivation of the upstream gradients to input point $\mathbf{x}_{n}$, which is generated by the backbone network:
\begin{align}
\label{eq:totalgrad2x}
    \dfrac{\partial \mathcal{E}}{\partial\mathbf{x}_{n}} & = \sum_{t=1}^{T}\dfrac{\partial \mathcal{E}}{\partial\mathbf{x}_{n}^{(t)}} \\
    & = 
    \label{eq:gradx_nstep_plus_gradx_estep}
    \sum_{t=1}^{T}\sum_{k=1}^{K}\underbrace{\dfrac{\partial\mathcal{E}}{\partial\boldsymbol{\mu}_{k}^{(t)}}\dfrac{\partial\boldsymbol{\mu}_{k}^{(t)}}{\partial\mathbf{x}_{n}}}_\textrm{grad. from N-step} + \underbrace{\dfrac{\partial\mathcal{E}}{\partial\boldsymbol{\mu}_{k}^{(t)}}\dfrac{\partial\boldsymbol{\mu}_{k}^{(t)}}{\partial\gamma_{nk}^{(t)}}{\dfrac{\partial\gamma_{nk}^{(t)}}{\partial\mathbf{x}_{n}}}}_\textrm{grad. from E-step} \\
    \label{eq:grad2x_approx}
    & \approx \sum_{t=1}^{T}\sum_{k=1}^{K}\left\{(1-\eta)^{(T-t)}\dfrac{\partial \mathcal{E}}{\partial \boldsymbol{\mu}_{k}^{(T)}}\right\}\left(\eta\dfrac{\gamma_{nk}^{(t)}}{N_{k}^{(t)}}\right)
\end{align}
 \Eref{eq:gradx_nstep_plus_gradx_estep} shows that, ignoring the gradients from the E-step, $\frac{\partial\mathcal{E}}{\partial\mathbf{x}_{n}}$ becomes a weighted sum of $\frac{\partial \boldsymbol{\mu}_{k}^{(t)}}{\partial \mathbf{x}_{n}}$ weighted by $\frac{\partial\mathcal{E}}{\partial\boldsymbol{\mu}_{k}^{(t)}}$, which is substituted with \Eref{eq:grad2mu}. As $\eta$ is reduced, the upstream gradients to earlier HEM layers grow relatively larger in magnitude, meaning that the loss gradients with respect to $\mathbf{x}_{n}$ become increasingly dominated by earlier HEM layers. Therefore, the backbone network can potentially learn better representations as it takes into account the effect of its parameters on not only the final HEM layers, but also on earlier HEM layers, where most of the convergence of the EM procedure occurs.
A full derivation is given in Appendix A.

\subsubsection{Forward-pass properties: improvement of ELBO}
In the forward pass, the N-step increases the ELBO for fractional step sizes, as shown in the following proposition:
\begin{prop}
For step size $\eta \in (0, 1]$, the N-step of a HEM iteration given by \Eref{eq:weighted_sum} updates $\boldsymbol{\mu}$ such that 
\begin{equation}
    \mathcal{L}(\boldsymbol{\mu}^{(t+1)}, \boldsymbol{\gamma}^{(t+1)}) > \mathcal{L}(\boldsymbol{\mu}^{(t)}, \boldsymbol{\gamma}^{(t+1)}),
\end{equation}
unless $\boldsymbol{\mu}^{(t+1)}=\boldsymbol{\mu}^{(t)}$, where $\mathcal{L}(\boldsymbol{\mu}, \boldsymbol{\gamma})$ is the ELBO defined for GMMs in \Eref{eq:elbo_gmm}, $\boldsymbol{\gamma}^{(t+1)}$ are the responsibilities computed in the E-step using $\boldsymbol{\mu}^{(t)}$, and $t$ is the iteration number.
\end{prop}

The proof is given in Appendix B. Since HEMNet is comprised of alternating E-steps and N-steps, it monotonically increases the ELBO, just as does the original unrolled EM. This is a non-trivial property of HEMNet. Monotonically increasing the ELBO converges the unrolled HEM iterations to a (local) maximum of the data-log likelihood. In other words, with every successive HEM iteration, the updated Gaussian bases $\boldsymbol{\mu}_{k}$ and attention weights $\gamma_{nk}$ can better reconstruct the original input points, as important semantics of the inputs are increasingly distilled into the GMM parameters \cite{li19emanet}.


\subsubsection{Computational Complexity}
\Eref{eq:weighted_sum} shows that computing the N-step update requires one additional operation to the original M-step: a convex summation of the M-step output $\mathcal{F}^{EM}\left(\boldsymbol{\mu}^{(t)}, \mathbf{X} \right)$ and the current $\boldsymbol{\mu}^{(t)}$ estimate. This operation incurs minimal additional computation and memory costs compared to the matrix multiplications in the E- and M-step, which dominates unrolled EM. Therefore, HEMNet has $\mathcal{O}(NK)$ space-time complexity, as does unrolled EM.

\section{Experiments}

\subsection{Implementation Details}
We use ResNet~\cite{resnet} pretrained on ImageNet~\cite{imagenet} with multi-grid~\cite{deeplabv3} as the backbone network. We use ResNet-50 with an output stride~(OS) of $16$ for all ablation studies and Resnet-101 with OS = $8$ for comparisons with other state-of-the-art approaches. We set the temperature $\sigma^{2}=\sqrt{C}$ following~\cite{transformer}, where $C$ is the number of input channels to HEMNet, which is set to $512$ by default. The step size is set to $\eta = 0.5$ for PASCAL VOC~\cite{pascal_voc} and PASCAL Context~\cite{pcontext}, and $\eta = 0.6$ for COCO Stuff~\cite{coco_stuff}. We set the training iteration number to $T_{train}=3$ for all datasets. We use the moving average mechanism~\cite{batchnorm, li19emanet} to update the initial Gaussian bases $\boldsymbol{\mu}^{(0)}$. We adopt the mean Intersection-over-Union~(mIoU) as the performance metric for all experiments across all datasets. Further details, including the training regime, are outlined in Appendix C.

\subsection{Ablation Studies}
\subsubsection{Ablation study on step size}
In Fig.~\ref{fig:grad2x}, we analyze the effects of the step size $\eta_{train}$ on the ELBO in the forward pass and the gradient flow in the backward pass. It can be seen that as $\eta_{train}$ is reduced, the convergence of the ELBO slows down, requiring more unrolled HEM layers than would otherwise (Fig.~\ref{fig:grad2x}a) . On the other hand, the gradients backpropagated to input $\mathbf{X}$ from each HEM layer, which is dominated by the N-step, become more evenly distributed as a greater proportion of upstream gradients are sent to earlier HEM layers (Fig.~\ref{fig:grad2x}b) . The subtlety here is that reducing $\eta_{train}$ does not seem to necessarily increase the magnitude of the total upstream gradients to $\mathbf{X}$ (Fig. \ref{fig:grad2x}d), since the gradients sent back to $\mathbf{X}$ from the $t$th HEM layer is proportional to $\eta(1-\eta)^{(T-t)}$, as shown in \Eref{eq:grad2x_approx}. This suggests that the change in performance from changing $\eta$ is likely due to the resulting change in relative weighting among the different HEM layers when computing the gradients with respect to $\mathbf{x}_{n}$, not the absolute magnitude of those gradients.

In other words, there is a trade-off, controlled by $\eta_{train}$, between the improvement of the ELBO in the forward pass and how evenly the early and later HEM layers contribute to backpropagating gradients to $\mathbf{X}$ in the backward pass. \Tref{table: ablation1} shows this trade-off, where the best performance for each value of $T_{train}$ is achieved by intermediate values of $\eta_{train}$, suggesting that they best manage this trade-off. Furthermore, the best $\eta_{train}$ decreases for larger $T_{train}$, as later HEM layers become increasingly redundant as they converge to an identity mapping of the GMM parameter estimates as the underyling EM procedure converges as well. Reducing the step size reduces the relative weighting on these redundant layers by increasing proportion of upstream gradients sent to earlier HEM layers, resulting in potentially better learned representations.
\begin{table}
\caption{Comparisons on PASCAL VOC val set in mIoU (\%). All results are computed for a ResNet-101 backbone, where OS = 8 for training and evaluation. FLOPs and memory are computed for input size of 513 $\times$ 513. \textbf{SS}: Single-scale input testing. \textbf{MS}: Multi-scale input testing. \textbf{Flip}: Adding left-right flipped input. (256), (512) denote the no. of input channels to EMANet and HEMNet.}
  \label{table:vocval}
  \centering
  \resizebox{\columnwidth}{!}{
  \begin{tabular}{lccccc}
    \toprule
    Method & SS & MS+Flip & FLOPs & Memory & Params\\
    \midrule
    ResNet-101 & - & - & 370.1G & 6.874G & 40.7M \\
    \midrule
    DeeplabV3+ \cite{deeplabv3+} & 77.62 & 78.72 & +142.6G & \underline{+318M} & +16.0M\\
    PSANet \cite{zhao2018psanet} & 78.51 & 79.77 & +185.7G & +528M & +26.4M\\
    EMANet (256) \cite{li19emanet} & 79.73 & 80.94 & \textbf{+45.2G} & \textbf{+236M} & \textbf{+5.15M} \\
    \textbf{HEMNet} (256) & \underline{80.93} & \underline{81.44} & \textbf{+45.2G} & \textbf{+236M} & \textbf{+5.15M} \\
    EMANet (512) \cite{li19emanet} & 80.05 & 81.32 & \underline{+92.3G} & +329M & \underline{+10.6M}\\
    \textbf{HEMNet} (512) & \textbf{81.33} & \textbf{82.23} & \underline{+92.3G} & +331M & \underline{+10.6M}\\
    \bottomrule
  \end{tabular}
  }
\end{table}
\begin{table}
  \caption{Comparisons on the PASCAL VOC test set.}
  \label{table:voc_test}
  \centering
  \resizebox{0.8\columnwidth}{!}{
  \begin{tabular}{l|c|c}
    \toprule
    Method & Backbone & mIoU (\%) \\
    \midrule
    PSPNet \cite{pspnet} & ResNet-101 & 85.4 \\
    DeeplabV3 \cite{deeplabv3} & ResNet-101 & 85.7 \\
    PSANet \cite{zhao2018psanet} & ResNet-101 & 85.7 \\
    EncNet \cite{encnet} & ResNet-101 & 85.9 \\
    DFN \cite{DFN} & ResNet-101 & 86.2\\
    Exfuse \cite{exfuse} & ResNet-101 & 86.2\\
    SDN \cite{SDN} & ResNet-101 & 86.6\\
    DIS \cite{DIS} & ResNet-101 & 86.8\\
    CFNet \cite{cfnet} & ResNet-101 & 87.2 \\
    EMANet \cite{li19emanet} & ResNet-101 & 87.7\\
    \textbf{HEMNet} & ResNet-101 & \textbf{88.0}\\
    \bottomrule
  \end{tabular}
  }
\end{table}
\subsubsection{Ablation study on iteration number}
We further investigate the effect of changing the training iteration number. It can be seen in Table \ref{table: ablation1} that for all step sizes $\eta_{train}$ the mIoU generally increases with the training iteration number $T_{train}$ up to a certain point, after which it decreases. This is likely attributed to the vanishing gradient problem as an exponentially less proportion of the upstream gradients reach earlier HEM layers as $T_{train}$ increases, meaning that the backbone network parameters are increasingly influenced by the later HEM layers, which amounts to a mere identity mapping of the GMM parameter estimates for high values of $T_{train}$. This is corroborated by the observation that the performance peaks at larger $T_{train}$ for smaller $\eta$, which can be explained by the fact that smaller $\eta$ slows down the exponential decay of the upstream gradients to earlier HEM layers, allowing us to unroll more HEM layers. In the case of EMANet \cite{li19emanet} the performance peaks at a low value of  $T_{train} = 2$, most likely because gradients aren't backpropagated through the unrolled EM iterations, preventing the rest of the network from learning representations optimized for the EM procedure.

Table \ref{table: ablation2} shows the effect of changing the evaluation iteration number, \textit{after} training. It can be seen that for all values of $\eta_{train}$ except $0.2$, increasing $T$ beyond $T_{train}$ during evaluation, where vanishing gradients is no longer an issue, can further improve performance. The observation can be attributed to the improvement in the ELBO with more HEM layers, which is consistent with previous findings \cite{jampani18ssn, li19emanet}. We suspect that the reason for the performance deterioration at high values of $T_{eval}$ is that the Gaussian bases have not fully converged at the chosen values of $T_{train}$ and that there is still room for the Gaussian bases' norms to change, making it difficult for HEMNet to generalize beyond its training horizon \cite{regularizing_rnns}.
\begin{table}
  \caption{Comparisons on the PASCAL Context test set. `+' means pretraining on COCO.}
  \label{table:pcontext}
  \centering
  \resizebox{0.8\columnwidth}{!}{
  \begin{tabular}{l|c|c}
    \toprule
    Method & Backbone & mIoU (\%) \\
    \midrule
    PSPNet \cite{pspnet} & ResNet-101 & 47.8 \\
    MSCI \cite{MSCI} & ResNet-152 & 50.3 \\
    SGR \cite{Liang2018SymbolicGR} & ResNet-101 & 50.8 \\
    CCL \cite{Ding2018ContextCF} & ResNet-101 & 51.6 \\
    EncNet \cite{encnet} & ResNet-101 & 51.7 \\
    SGR+ \cite{Liang2018SymbolicGR} & ResNet-101 & 52.5 \\
    DANet \cite{danet} & ResNet-101 & 52.6 \\
    EMANet \cite{li19emanet} & ResNet-101 & 53.1\\
    CFNet \cite{cfnet} & ResNet-101 & 54.0 \\
    \textbf{HEMNet} & ResNet-101 & \textbf{54.3} \\
    \bottomrule
  \end{tabular}
  }
\end{table}
\begin{table}
\caption{Comparisons on COCO Stuff test set.}
    \label{table:cocostufftest}
    \centering
    \resizebox{\columnwidth}{!}{
    \begin{tabular}{l|c|c}
        \toprule
        Method & Backbone & mIoU (\%) \\
        \midrule
        RefineNet \cite{refinenet} & ResNet-101 & 33.6 \\
        CCL \cite{Ding2018ContextCF} & ResNet-101 & 35.7 \\
        DSSPN \cite{Liang2018DynamicStructuredSP} & ResNet-101 & 37.3 \\
        SGR \cite{Liang2018SymbolicGR} & ResNet-101 & 39.1 \\
        DANet \cite{danet} & ResNet-101 & 39.7 \\
        EMANet \cite{li19emanet} & ResNet-101 & 39.9 \\
        \textbf{HEMNet} & ResNet-101 & \textbf{40.1} \\ 
        \bottomrule
    \end{tabular}
    }
\end{table}

\begin{figure*}[t]
    \captionsetup[subfigure]{labelformat=empty}
    \centering
    \begin{rotate}{90}\footnotesize{}\end{rotate}
    \begin{subfigure}[b]{0.40\linewidth}
        \centering
        \caption{Image \qquad\quad\quad Label \qquad\quad\quad Prediction}
        \vspace{-2mm}
        \includegraphics[width=\linewidth]{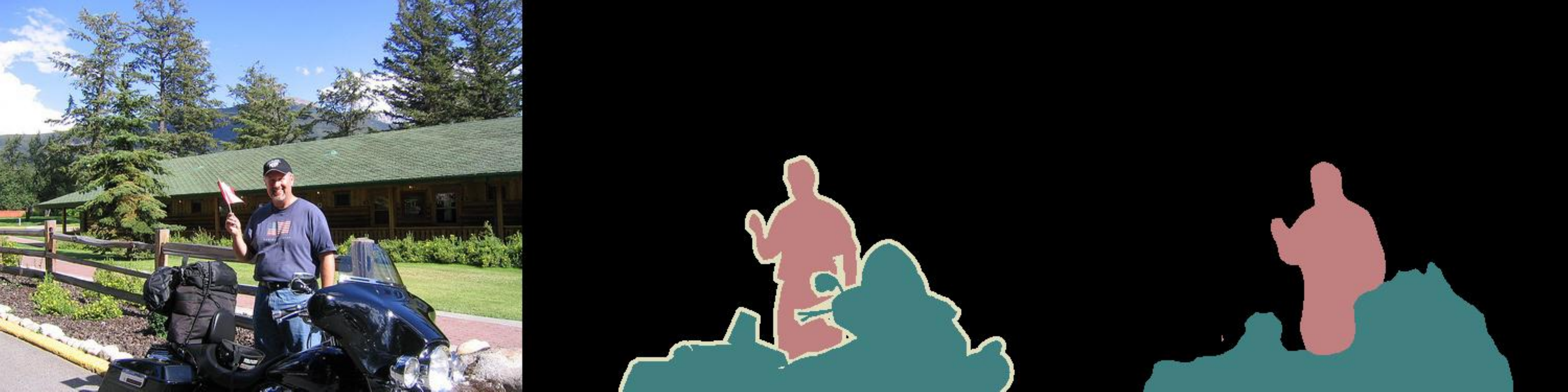}
    \end{subfigure}
    \begin{subfigure}[b]{0.40\linewidth}
        \centering
        \caption{Input Feature Maps $\mathbf{X}$}
        \vspace{-2mm}
        \includegraphics[width=\linewidth]{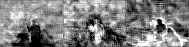}
    \end{subfigure}\\
    \vspace{2mm}
    \begin{rotate}{90}\footnotesize{~\quad\quad\quad $\boldsymbol{t} = 3$\quad\quad\quad~ $\boldsymbol{t} = 2$\quad\quad\quad~ $\boldsymbol{t} = 1$}\end{rotate}
    \begin{subfigure}[b]{0.40\linewidth}
        \centering
        \caption{Attention Maps $\boldsymbol{\gamma}^{(t)}$}
        \vspace{-2mm}
        \includegraphics[width=\linewidth]{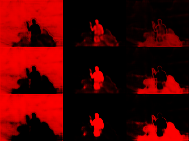}
        \caption{$\boldsymbol{\gamma}_{\cdot i}$\qquad\qquad\qquad$\boldsymbol{\gamma}_{\cdot j}$\qquad\qquad\qquad$\boldsymbol{\gamma}_{\cdot k}$}
    \end{subfigure}
    \begin{subfigure}[b]{0.40\linewidth}
        \centering
        \caption{Reconstructed Inputs $\widetilde{\mathbf{X}}^{(t)}$}
        \vspace{-2mm}
        \includegraphics[width=\linewidth]{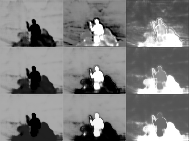}
        \caption{$\widetilde{\mathbf{X}}_{\cdot a}$\qquad\qquad\qquad$\widetilde{\mathbf{X}}_{\cdot b}$\qquad\qquad\qquad$\widetilde{\mathbf{X}}_{\cdot c}$}
    \end{subfigure}
    \caption{Visualization of the attention maps $\boldsymbol{\gamma}$, input feature maps $\mathbf{X}$ from the backbone CNN, and reconstructed inputs $\widetilde{\mathbf{X}}=\boldsymbol{\gamma\mu}$ for a sample image from the PASCAL VOC 2012 val set. The images in the top-left corner contain the original image, label, and the prediction made by HEMNet with a ResNet-101 backbone. The three rows below show the attention maps (left) and reconstructed inputs (right) at each HEM iteration. $\boldsymbol{\gamma}_{\cdot k}$ denotes the attention map w.r.t. the $k$th Gaussian basis $\boldsymbol{\mu}_{k}$ and $\widetilde{\mathbf{X}}_{\cdot c}$ denotes the $c$th channel of the reconstructed input, where $1\leq i,j,k\leq K$ and $1\leq a,b,c \leq C$.}
    \label{fig:viz}
\end{figure*}
\subsection{Comparisons with State-of-the-arts}
We first compare our approach to EMANet \cite{li19emanet} and other baselines on the PASCAL VOC validation set. \Tref{table:vocval} shows that HEMNet outperforms all baselines by a large margin. Most notably, HEMNet outperforms EMANet while incurring virtually the same computation and memory costs and using the same number of parameters. Furthermore, HEMNet with 256 input channels exceeds EMANet with 512 input channels, and HEMNet's single-scale (SS) performance is on par with EMANet's \textit{multi-scale} (MS) performance, which is a robust method for improving semantic segmentation accuracy \cite{pspnet, deeplabv3+, danet, li19emanet}. We also display significant improvement in performance over EMANet on the PASCAL VOC, PASCAL Context and COCO Stuff test set, as shown in \Tref{table:voc_test}, \Tref{table:pcontext}, and \Tref{table:cocostufftest}.

\subsection{Visualizations}
In Fig.~\ref{fig:viz}, we demonstrate the effect of improving the ELBO in HEMNet by visualizing the responsibilities $\boldsymbol{\gamma}$, which act as attention weights, and the feature map $\widetilde{\mathbf{X}}$ reconstructed from those attention weights. It can be seen that the attention map with respect to each Gaussian basis attends to a specific aspect of the input image, suggesting that each Gaussian basis corresponds to a particular semantic concept. For instance, the attention maps with respect to bases $i$, $j$, and $k$ appear to highlight the background, person and motorbike, respectively. 

Furthermore, after every HEM iteration the attention maps grow sharper and converges to the underlying semantics of each basis. As a result, every feature map reconstructed from the updated $\boldsymbol{\gamma}$ and $\boldsymbol{\mu}$ increasingly recovers its fundamental semantics from the noisy input $\mathbf{X}$, while progressively suppressing irrelevant concepts and details. This stems from the fact that every HEM iteration monotonically increases the ELBO, and hence the log-likelihood with respect to the input features, thereby removing unnecessary noise and distilling the important semantics from $\mathbf{X}$ into Gaussian bases $\boldsymbol{\mu}$.

\section{Conclusion}
In this work, we proposed Highway Expectation Maximization Networks~(HEMNet) in order to address the vanishing gradient problem present in expectation maximization~(EM) iterations. The proposed HEMNet is comprised of unrolled iterations of the generalized EM algorithm based on the Newton-Rahpson method, which introduces skip connections that ameliorate gradient flow while preserving the underlying EM procedure and incurring minimal additional computation and memory costs. We performed extensive ablation studies and experiments on several semantic segmentation benchmarks to demonstrate that our approach effectively alleviates vanishing gradients and enables better performance.

\section*{Broader Impact}

Our research can be broadly described as an attempt to leverage what has been two very successful approaches to machine learning: model-based methods and deep neural networks. Deep learning methods have shown state-of-the-art performance in a variety of applications due to their excellent ability for representation learning, but they are considered as black box methods, making it extremely difficult to interpret their inner workings or incorporate domain knowledge of the problem at hand. Combining deep neural networks with well-studied, classical model-based approaches provide a straightforward way to incorporate problem specific assumptions, such as those from the physical world like three-dimensional geometry or visual occlusion. Furthermore, this combined approach also provides a level of interpretability to the inner workings of the system, through the lens of well-studied classical methods. In our paper we consider a semantic segmentation task, which has become a key component of the vision stack in autonomous driving technology. Combining deep neural networks with the model-based approaches can be crucial for designing and analyzing the potential failure modes of such safety-critical applications of deep-learning based systems. For instance in our approach, the Gaussian bases converge to specific semantics with every passing EM iteration, and the corresponding attention maps give an indication of the underlying semantics of a given input image. This can help researchers better understand what the system is actually learning, which can be invaluable, for example, when trying to narrow down the reason for an accident induced by a self-driving vehicle. We hope that our work encourages researchers to incorporate components inspired by classical model-based machine learning into existing deep learning architectures, enabling them to embed their domain knowledge into the design of the network, which in turn may provide another means to better interpret its inner mechanism.

\bibliography{references}

\begin{thebibliography}{46}
\providecommand{\natexlab}[1]{#1}
\providecommand{\url}[1]{\texttt{#1}}
\providecommand{\urlprefix}{URL }
\expandafter\ifx\csname urlstyle\endcsname\relax
  \providecommand{\doi}[1]{doi:\discretionary{}{}{}#1}\else
  \providecommand{\doi}{doi:\discretionary{}{}{}\begingroup
  \urlstyle{rm}\Url}\fi

\bibitem[{Bapna et~al.(2018)Bapna, Chen, Firat, Cao, and Wu}]{transparentatt}
Bapna, A.; Chen, M.; Firat, O.; Cao, Y.; and Wu, Y. 2018.
\newblock Training Deeper Neural Machine Translation Models with Transparent
  Attention.
\newblock In \emph{Proceedings of the 2018 Conference on Empirical Methods in
  Natural Language Processing}, 3028--3033.

\bibitem[{Bengio, Simard, and Frasconi(1994)}]{vanishinggradients}
Bengio, Y.; Simard, P.; and Frasconi, P. 1994.
\newblock Learning long-term dependencies with gradient descent is difficult.
\newblock \emph{IEEE Transactions on Neural Networks} 5(2): 157--166.

\bibitem[{Caesar, Uijlings, and Ferrari(2018)}]{coco_stuff}
Caesar, H.; Uijlings, J.; and Ferrari, V. 2018.
\newblock COCO-Stuff: Thing and Stuff Classes in Context.
\newblock In \emph{IEEE Conference on Computer Vision and Pattern Recognition
  (CVPR)}, 1209--1218.

\bibitem[{Chen et~al.(2017)Chen, Papandreou, Schroff, and Adam}]{deeplabv3}
Chen, L.-C.; Papandreou, G.; Schroff, F.; and Adam, H. 2017.
\newblock Rethinking atrous convolution for semantic image segmentation.
\newblock \emph{arXiv preprint arXiv:1706.05587} .

\bibitem[{Chen et~al.(2018)Chen, Zhu, Papandreou, Schroff, and
  Adam}]{deeplabv3+}
Chen, L.-C.; Zhu, Y.; Papandreou, G.; Schroff, F.; and Adam, H. 2018.
\newblock Encoder-Decoder with Atrous Separable Convolution for Semantic Image
  Segmentation.
\newblock In \emph{European Conference on Computer Vision (ECCV)}, 801--818.

\bibitem[{Cho et~al.(2014)Cho, van Merri{\"e}nboer, Gulcehre, Bahdanau,
  Bougares, Schwenk, and Bengio}]{gru}
Cho, K.; van Merri{\"e}nboer, B.; Gulcehre, C.; Bahdanau, D.; Bougares, F.;
  Schwenk, H.; and Bengio, Y. 2014.
\newblock Learning Phrase Representations using {RNN} Encoder{--}Decoder for
  Statistical Machine Translation.
\newblock In \emph{Proceedings of the 2014 Conference on Empirical Methods in
  Natural Language Processing ({EMNLP})}, 1724--1734.

\bibitem[{David~{Krueger}(2016)}]{regularizing_rnns}
David~{Krueger}, R.~M. 2016.
\newblock Regularizing RNNs by Stabilizing Activations.
\newblock In \emph{4th International Conference on Learning Representations,
  {ICLR} 2016, San Juan, Puerto Rico, May 2-4, 2016, Conference Track
  Proceedings}.

\bibitem[{Dempster, Laird, and Rubin(1977)}]{Dempster77maximumlikelihood}
Dempster, A.~P.; Laird, N.~M.; and Rubin, D.~B. 1977.
\newblock Maximum likelihood from incomplete data via the EM algorithm.
\newblock \emph{Journal of the Royal Statistical Society, Series B} 39(1):
  1--38.

\bibitem[{Ding et~al.(2018)Ding, Jiang, Shuai, Liu, and
  Wang}]{Ding2018ContextCF}
Ding, H.; Jiang, X.; Shuai, B.; Liu, A.~Q.; and Wang, G. 2018.
\newblock Context Contrasted Feature and Gated Multi-scale Aggregation for
  Scene Segmentation.
\newblock \emph{{IEEE} Conference on Computer Vision and Pattern Recognition
  (CVPR)} 2393--2402.

\bibitem[{Everingham et~al.(2010)Everingham, Gool, Williams, Winn, and
  Zisserman}]{pascal_voc}
Everingham, M.; Gool, L.; Williams, C.~K.; Winn, J.; and Zisserman, A. 2010.
\newblock The Pascal Visual Object Classes (VOC) Challenge.
\newblock \emph{Int. J. Comput. Vision} 88(2): 303–338.

\bibitem[{Fu et~al.(2019)Fu, Liu, Tian, Li, Bao, Fang, and Lu}]{danet}
Fu, J.; Liu, J.; Tian, H.; Li, Y.; Bao, Y.; Fang, Z.; and Lu, H. 2019.
\newblock Dual Attention Network for Scene Segmentation.
\newblock In \emph{{IEEE} Conference on Computer Vision and Pattern Recognition
  (CVPR)}, 3146--3154.

\bibitem[{{Fu} et~al.(2019){Fu}, {Liu}, {Wang}, {Zhou}, {Wang}, and {Lu}}]{SDN}
{Fu}, J.; {Liu}, J.; {Wang}, Y.; {Zhou}, J.; {Wang}, C.; and {Lu}, H. 2019.
\newblock Stacked Deconvolutional Network for Semantic Segmentation.
\newblock \emph{IEEE Transactions on Image Processing} 1--1.

\bibitem[{Gers, Schmidhuber, and Cummins(1999)}]{lstm_with_forget_gate}
Gers, F.~A.; Schmidhuber, J.; and Cummins, F. 1999.
\newblock Learning to forget: continual prediction with LSTM.
\newblock In \emph{1999 Ninth International Conference on Artificial Neural
  Networks ICANN 99. (Conf. Publ. No. 470)}, volume~2, 850--855 vol.2.

\bibitem[{Greff, Van~Steenkiste, and Schmidhuber(2017)}]{nem}
Greff, K.; Van~Steenkiste, S.; and Schmidhuber, J. 2017.
\newblock Neural expectation maximization.
\newblock In \emph{Advances in Neural Information Processing Systems (NIPS)},
  6691--6701.

\bibitem[{He et~al.(2016{\natexlab{a}})He, Zhang, Ren, and Sun}]{resnet}
He, K.; Zhang, X.; Ren, S.; and Sun, J. 2016{\natexlab{a}}.
\newblock Deep residual learning for image recognition.
\newblock In \emph{{IEEE} Conference on Computer Vision and Pattern Recognition
  (CVPR)}, 770--778.

\bibitem[{He et~al.(2016{\natexlab{b}})He, Zhang, Ren, and
  Sun}]{identity_mappings_in_resnets}
He, K.; Zhang, X.; Ren, S.; and Sun, J. 2016{\natexlab{b}}.
\newblock Identity Mappings in Deep Residual Networks.
\newblock In \emph{European Conference on Computer Vision (ECCV)}, 630--645.

\bibitem[{Hershey, Roux, and Weninger(2014)}]{deep_unfolding}
Hershey, J.~R.; Roux, J.~L.; and Weninger, F. 2014.
\newblock Deep Unfolding: Model-Based Inspiration of Novel Deep Architectures.
\newblock \emph{arXiv preprint arXiv:1409.2574} .

\bibitem[{Hinton, Sabour, and Frosst(2018)}]{emrouting}
Hinton, G.~E.; Sabour, S.; and Frosst, N. 2018.
\newblock Matrix capsules with {EM} routing.
\newblock In \emph{International Conference on Learning Representations}.

\bibitem[{Hochreiter and Schmidhuber(1997)}]{lstm}
Hochreiter, S.; and Schmidhuber, J. 1997.
\newblock Long short-term memory.
\newblock \emph{Neural computation} 9(8): 1735--1780.

\bibitem[{Huang et~al.(2017)Huang, Liu, Van Der~Maaten, and
  Weinberger}]{densenet}
Huang, G.; Liu, Z.; Van Der~Maaten, L.; and Weinberger, K.~Q. 2017.
\newblock Densely connected convolutional networks.
\newblock In \emph{{IEEE} Conference on Computer Vision and Pattern Recognition
  (CVPR)}, 4700--4708.

\bibitem[{Ioffe and Szegedy(2015)}]{batchnorm}
Ioffe, S.; and Szegedy, C. 2015.
\newblock Batch Normalization: Accelerating Deep Network Training by Reducing
  Internal Covariate Shift.
\newblock In \emph{Proceedings of the 32nd International Conference on Machine
  Learning}, 448--456.

\bibitem[{Jampani et~al.(2018)Jampani, Sun, Liu, Yang, and
  Kautz}]{jampani18ssn}
Jampani, V.; Sun, D.; Liu, M.-Y.; Yang, M.-H.; and Kautz, J. 2018.
\newblock Superpixel sampling networks.
\newblock In \emph{European Conference on Computer Vision (ECCV)}, 352--368.

\bibitem[{Lange(1995)}]{gem_newton}
Lange, K. 1995.
\newblock A gradient algorithm locally equivalent to the EM algorithm.
\newblock \emph{Journal of the Royal Statistical Society: Series B
  (Methodological)} 57(2): 425--437.

\bibitem[{Li et~al.(2019)Li, Zhong, Wu, Yang, Lin, and Liu}]{li19emanet}
Li, X.; Zhong, Z.; Wu, J.; Yang, Y.; Lin, Z.; and Liu, H. 2019.
\newblock Expectation-Maximization Attention Networks for Semantic
  Segmentation.
\newblock In \emph{{IEEE} International Conference on Computer Vision (ICCV)},
  9167--9176.

\bibitem[{Liang et~al.(2018)Liang, Hu, Zhang, Lin, and
  Xing}]{Liang2018SymbolicGR}
Liang, X.; Hu, Z.; Zhang, H.; Lin, L.; and Xing, E.~P. 2018.
\newblock Symbolic graph reasoning meets convolutions.
\newblock In \emph{Advances in Neural Information Processing Systems (NIPS)},
  1853--1863.

\bibitem[{Liang, Zhou, and Xing(2018)}]{Liang2018DynamicStructuredSP}
Liang, X.; Zhou, H.; and Xing, E. 2018.
\newblock Dynamic-Structured Semantic Propagation Network.
\newblock \emph{{IEEE} Conference on Computer Vision and Pattern Recognition
  (CVPR)} 752--761.

\bibitem[{Lin et~al.(2018)Lin, Ji, Lischinski, Cohen-Or, and Huang}]{MSCI}
Lin, D.; Ji, Y.; Lischinski, D.; Cohen-Or, D.; and Huang, H. 2018.
\newblock Multi-scale context intertwining for semantic segmentation.
\newblock In \emph{European Conference on Computer Vision (ECCV)}, 603--619.

\bibitem[{Lin et~al.(2017)Lin, Milan, Shen, and Reid}]{refinenet}
Lin, G.; Milan, A.; Shen, C.; and Reid, I. 2017.
\newblock Refine{N}et: {M}ulti-Path Refinement Networks for High-Resolution
  Semantic Segmentation.
\newblock In \emph{{IEEE} Conference on Computer Vision and Pattern Recognition
  (CVPR)}, 1925--1934.

\bibitem[{{Luo} et~al.(2017){Luo}, {Wang}, {Lin}, and {Wang}}]{DIS}
{Luo}, P.; {Wang}, G.; {Lin}, L.; and {Wang}, X. 2017.
\newblock Deep Dual Learning for Semantic Image Segmentation.
\newblock In \emph{2017 IEEE International Conference on Computer Vision
  (ICCV)}, 2737--2745.

\bibitem[{Mottaghi et~al.(2014)Mottaghi, Chen, Liu, Cho, Lee, Fidler, Urtasun,
  and Yuille}]{pcontext}
Mottaghi, R.; Chen, X.; Liu, X.; Cho, N.-G.; Lee, S.-W.; Fidler, S.; Urtasun,
  R.; and Yuille, A. 2014.
\newblock The Role of Context for Object Detection and Semantic Segmentation in
  the Wild.
\newblock In \emph{IEEE Conference on Computer Vision and Pattern Recognition
  (CVPR)}, 891–898.

\bibitem[{Neal and Hinton(1999)}]{coordinatedescent}
Neal, R.~M.; and Hinton, G.~E. 1999.
\newblock \emph{A View of the EM Algorithm That Justifies Incremental, Sparse,
  and Other Variants}, 355–368.
\newblock Cambridge, MA, USA: MIT Press.

\bibitem[{Richardson and Green(1997)}]{gmm}
Richardson, S.; and Green, P.~J. 1997.
\newblock On Bayesian analysis of mixtures with an unknown number of components
  (with discussion).
\newblock \emph{Journal of the Royal Statistical Society: series B (statistical
  methodology)} 59(4): 731--792.

\bibitem[{Russakovsky et~al.(2015)Russakovsky, Deng, Su, Krause, Satheesh, Ma,
  Huang, Karpathy, Khosla, Bernstein, Berg, and Fei-Fei}]{imagenet}
Russakovsky, O.; Deng, J.; Su, H.; Krause, J.; Satheesh, S.; Ma, S.; Huang, Z.;
  Karpathy, A.; Khosla, A.; Bernstein, M.; Berg, A.~C.; and Fei-Fei, L. 2015.
\newblock {ImageNet Large Scale Visual Recognition Challenge}.
\newblock \emph{International Journal of Computer Vision (IJCV)} 115(3):
  211--252.

\bibitem[{Srivastava, Greff, and Schmidhuber(2015)}]{highway_nets}
Srivastava, R.~K.; Greff, K.; and Schmidhuber, J. 2015.
\newblock Training very deep networks.
\newblock In \emph{Advances in Neural Information Processing Systems (NIPS)},
  2377--2385.

\bibitem[{Vaswani et~al.(2017)Vaswani, Shazeer, Parmar, Uszkoreit, Jones,
  Gomez, Kaiser, and Polosukhin}]{transformer}
Vaswani, A.; Shazeer, N.; Parmar, N.; Uszkoreit, J.; Jones, L.; Gomez, A.~N.;
  Kaiser, {\L}.; and Polosukhin, I. 2017.
\newblock Attention is all you need.
\newblock In \emph{Advances in Neural Information Processing Systems (NIPS)},
  5998--6008.

\bibitem[{Wang et~al.(2019)Wang, Li, Xiao, Zhu, Li, Wong, and
  Chao}]{deeptransformer}
Wang, Q.; Li, B.; Xiao, T.; Zhu, J.; Li, C.; Wong, D.~F.; and Chao, L.~S. 2019.
\newblock Learning deep transformer models for machine translation.
\newblock \emph{arXiv preprint arXiv:1906.01787} .

\bibitem[{Wang et~al.(2018)Wang, Girshick, Gupta, and He}]{NonLocal2018}
Wang, X.; Girshick, R.; Gupta, A.; and He, K. 2018.
\newblock Non-local Neural Networks.
\newblock In \emph{{IEEE} Conference on Computer Vision and Pattern Recognition
  (CVPR)}, 7794--7803.

\bibitem[{Wang et~al.(2020)Wang, Wang, Yang, Li, Li, and Li}]{wang_2020}
Wang, Z.; Wang, S.; Yang, S.; Li, H.; Li, J.; and Li, Z. 2020.
\newblock Weakly Supervised Fine-Grained Image Classification via Guassian
  Mixture Model Oriented Discriminative Learning.
\newblock In \emph{Proceedings of the IEEE/CVF Conference on Computer Vision
  and Pattern Recognition (CVPR)}.

\bibitem[{Wu(1983)}]{wu1983}
Wu, C.~J. 1983.
\newblock On the Convergence Properties of the EM Algorithm.
\newblock \emph{The Annals of statistics} 95--103.

\bibitem[{Yu et~al.(2018)Yu, Wang, Peng, Gao, Yu, and Sang}]{DFN}
Yu, C.; Wang, J.; Peng, C.; Gao, C.; Yu, G.; and Sang, N. 2018.
\newblock Learning a Discriminative Feature Network for Semantic Segmentation.
\newblock In \emph{The IEEE Conference on Computer Vision and Pattern
  Recognition (CVPR)}.

\bibitem[{Zhang, Titov, and Sennrich(2019)}]{improving_deep_transformer}
Zhang, B.; Titov, I.; and Sennrich, R. 2019.
\newblock Improving Deep Transformer with Depth-Scaled Initialization and
  Merged Attention.
\newblock In \emph{Proceedings of the 2019 Conference on Empirical Methods in
  Natural Language Processing and the 9th International Joint Conference on
  Natural Language Processing (EMNLP-IJCNLP)}, 898--909.

\bibitem[{Zhang et~al.(2018{\natexlab{a}})Zhang, Dana, Shi, Zhang, Wang, Tyagi,
  and Agrawal}]{encnet}
Zhang, H.; Dana, K.; Shi, J.; Zhang, Z.; Wang, X.; Tyagi, A.; and Agrawal, A.
  2018{\natexlab{a}}.
\newblock Context encoding for semantic segmentation.
\newblock In \emph{{IEEE} Conference on Computer Vision and Pattern Recognition
  (CVPR)}, 7151--7160.

\bibitem[{Zhang et~al.(2019)Zhang, Zhang, Wang, and Xie}]{cfnet}
Zhang, H.; Zhang, H.; Wang, C.; and Xie, J. 2019.
\newblock Co-Occurrent Features in Semantic Segmentation.
\newblock In \emph{{IEEE} Conference on Computer Vision and Pattern Recognition
  (CVPR)}, 548--557.

\bibitem[{Zhang et~al.(2018{\natexlab{b}})Zhang, Zhang, Peng, Xue, and
  Sun}]{exfuse}
Zhang, Z.; Zhang, X.; Peng, C.; Xue, X.; and Sun, J. 2018{\natexlab{b}}.
\newblock ExFuse: Enhancing Feature Fusion for Semantic Segmentation.
\newblock In \emph{Proceedings of the European Conference on Computer Vision
  (ECCV)}.

\bibitem[{Zhao et~al.(2017)Zhao, Shi, Qi, Wang, and Jia}]{pspnet}
Zhao, H.; Shi, J.; Qi, X.; Wang, X.; and Jia, J. 2017.
\newblock Pyramid scene parsing network.
\newblock In \emph{{IEEE} Conference on Computer Vision and Pattern Recognition
  (CVPR)}, 2881--2890.

\bibitem[{Zhao et~al.(2018)Zhao, Zhang, Liu, Shi, Loy, Lin, and
  Jia}]{zhao2018psanet}
Zhao, H.; Zhang, Y.; Liu, S.; Shi, J.; Loy, C.~C.; Lin, D.; and Jia, J. 2018.
\newblock {PSANet}: Point-wise Spatial Attention Network for Scene Parsing.
\newblock In \emph{European Conference on Computer Vision (ECCV)}, 267--283.

\end{thebibliography}

\end{document}